\documentclass{article} 
\usepackage{nips13submit_e,times}
\usepackage{hyperref}
\usepackage{url}

\usepackage{amsmath}
\usepackage{algorithmic}
\usepackage{algorithm}
\usepackage[]{graphicx}

\title{Modeling correlations in spontaneous activity of visual cortex with centered
Gaussian-binary deep Boltzmann machines}

\author{
Nan Wang\\
Institut f\"ur Neuroinformatik\\
Ruhr-Universit\"at Bochum\\
Bochum, 44780, Germany\\
\texttt{nan.wang@ini.rub.de} \\
\And
Dirk Jancke\\
Institut f\"ur Neuroinformatik\\
Ruhr-Universit\"at Bochum\\
Bochum, 44780, Germany\\
\texttt{jancke@neurobiologie.rub.de} \\
\And
Laurenz Wiskott\\
Institut f\"ur Neuroinformatik\\
Ruhr-Universit\"at Bochum\\
Bochum, 44780, Germany\\
\texttt{laurenz.wiskott@rub.de} \\
}

\newcommand{\vect}[1]{\mathbf{#1}}
\nipsfinalcopy 

\begin{document}

\maketitle

\begin{abstract}
Spontaneous cortical activity -- the ongoing cortical activities in absence of
intentional sensory input -- is considered to
play a vital role in many aspects of both normal brain
functions~\cite{KenetBibitchkovEtAl-2003} and mental
dysfunctions~\cite{Burke-2002}. 
We present a centered Gaussian-binary Deep Boltzmann Machine (GDBM) for
modeling the spontaneous activity in early cortical visual area and relate the
random sampling in GDBMs to the
spontaneous cortical activity. After training the proposed model on natural
image patches, we show that the samples collected from the model's probability distribution
encompass similar activity patterns as found in the spontaneous activity.
Specifically, filters having the same orientation
preference tend to be active together during random sampling. Our
work demonstrates GDBM is a meaningful model approach for basic receptive field
properties and the emergence of spontaneous activity patterns in early cortical
visual areas. Besides, we show empirically that centered GDBMs
do not suffer from the difficulties during training as GDBMs do and can be
properly trained without the layer-wise pretraining as described in~\cite{ChoRaikoEtAl-2013}.
\end{abstract}

\section{Introduction}
Spontaneous cortical activity has been studied in various contexts
ranging from somato-sensory to visual and auditory
perception~\cite{Lestienne-2001}. 
These ongoing cortical activities in absence of intentional sensory input are considered to
play a vital role in many aspects of both normal brain
functions~\cite{KenetBibitchkovEtAl-2003}, and mental
dysfunctions~\cite{Burke-2002}.  
In~\cite{KenetBibitchkovEtAl-2003}, the spontaneous activity in the early cortical visual
area is reported to have a set of states, several of which resemble cortical
representation of orientation, i.e.\ neurons in the
visual cortex having similar orientation preference tend to be active together.
One hypothesis about this finding is that these states reflect expectations
about the sensory input, which reflects an aspect of approaches such as Bayesian
models\cite{VilaresKording-2011}  and generative models. Concretely, a brain
learn to synthesize (or generate) representations of sensory
inputs. 

Previous studies have considered Deep Boltzmann Machines (DBMs)
as a potential model framework for modeling the relevant aspects of 
this generative process and have related the inference in DBM to the
mechanisms of cortical perception\cite{ReichertSeriesEtAl-2010,ReichertSeriesEtAl-2013}.
The authors have trained a DBM on binary images and have shown
that trained DBMs can qualitatively reproduce several aspects of hallucinations
in Charles Bonnet Syndrome.
In this work, we have chosen a variant of DBM, a Gaussian-Binary DBM (GDBM), as
a model for early cortical visual areas, which extends
DBMs to modeling continuous data. In particular, we are interested in modeling
the spontaneous activity found in~\cite{KenetBibitchkovEtAl-2003} and
demonstrate how GDBM reproduces the findings of spontaneous activity
in early visual cortex. 

As for training GDBMs on the visual inputs, we adapt the centering
proposed in~\cite{MontavonMueller-2012} to GDBMs by rewriting the energy function as
centered states. We show empirically the proposed centered GDBMs can learn
features from the natural image patches without the layer-wise pretraining
procedure and do not suffer from the difficulties during training of GDBMs as reported
in~\cite{ChoRaikoEtAl-2013}.

In Section~\ref{sec:algorithm}, the proposed centered GDBM is introduced. Then
we describe the training procedure in Section~\ref{sec:naturalImages}, and show
that the centered GDBM can not only learn Gabor-like filters, but also
meaningful features in the higher layer. Finally, by considering the
samples of the hidden units from the model distribution as spontaneous
activity, we demonstrate in Section~\ref{sec:spontaneousExp} that these
random samples present similar activity patterns as the spontaneous visual cortical
activity reported in~\cite{KenetBibitchkovEtAl-2003}.  

\section{Algorithm}
\label{sec:algorithm}
Like Deep Belief Networks (DBN)~\cite{HintonSalakhutdinov-2006} and deep
autoencoders~\cite{VincentLarochelleEtAl-2008}, DBMs have been proposed for
learning multi-layer representations that are increasingly complex. In
particular, DBM incorporates both the
top-down messages and the bottom-up passes during the inference for each layer,
which gives DBM an advantage in propagating the input uncertainties. DBMs have
been applied to many problems and show promising
results\cite{SrivastavaSalakhutdinov-2012, SrivastavaSalakhutdinovEtAl-2013,
SalakhutdinovTenenbaumEtAl-2013}. 

To model the natural image patches that have continuous values, a variant of
DBMs is required since the original DBM is designed for binary data. There are
two common ways to extend DBMs to modeling continuous values. The most common way
is to train a Gaussian-binary restricted Boltzmann machine (GRBM) as a
preprocessing model and use the output of the trained model as the input data
for training a DBM~\cite{SalakhutdinovHinton-2009}. However, this practice
loses the ability to train the model
as a whole, i.e.\ the preprocessing part needs to be trained beforehand. An
alternative is a Gaussian-binary deep Boltzmann machine (also known as
Gaussian-Bernoulli DBM~\cite{ChoRaikoEtAl-2013}), in which the binary units in
the bottom layer are replaced by the real-value ones as in GRBMs. Moreover,
GDBMs have been proved to be a universal approximator. The pitfall of GDBMs is the
difficulty in training it~\cite{ChoRaikoEtAl-2013}. 

The centering has been proposed in~\cite{MontavonMueller-2012} for DBMs
and has been shown to improve learning. The idea is to have the output of each
unit to all the other units to be centered around zero and take effects only when
its activation deviate from its mean activation.
In~\cite{MontavonMueller-2012}, the authors show
that centering produces a better conditioned optimization problem. 
Therefore, we adapted the centering to the GDBM and refer to the new model as centered GDBM.
Compared with the training recipe in~\cite{ChoRaikoEtAl-2013}, we find
empirically that a centered GDBM can easily be trained even without
pre-training phase and is insensitive to the choice of hyper-parameters.
	\subsection{Centered GDBM}
\begin{figure}[ht]
		\centering
		\includegraphics[width=0.7\linewidth]{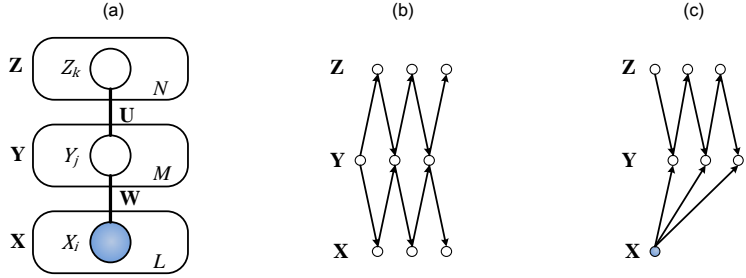}
		\caption{(a) A graphical description of a two-layer GDBM. (b) Illustration of the
		sampling procedure for estimating model dependent expectation.
		(c) Illustration of the sampling procedure for estimating data dependent expectation.
		}
		\label{fig:gdbm}
	\end{figure}

	To avoid cluttering, we present a centered GDBM of two hidden layers as
	an example, although the model can be extended to an arbitrary number
	of layers. A two-layer centered GDBM as illustrated in
	Figure~\ref{fig:gdbm}, consisted of an input layer
	$\vect{X}$ and two hidden layers, has an energy defined as:
	\begin{eqnarray}
	E\left( \mathbf{X}, \mathbf{Y} , \mathbf{Z}; \Theta, \mathcal{C}\right):&=&
	\sum_i^M \frac{\left( X_i-c_{x_i} \right)^2}{2\sigma_i^2}
	- \sum_{i,j}^{L,M} \frac{ (X_i-c_{x_i}) w_{ij} (Y_j-c_{y_j}) }{\sigma_i^2} 
	- \sum_j^M b_{y_j} (Y_j - c_{y_j}) 
	\nonumber \\
	&&
	- \sum_k^N b_{z_k} (Z_k - c_{z_k})
	- \sum_{j,k}^{M,N} (Y_j - c_{y_j}) u_{jk} (Z_k - c_{z_k}) 
	\\
	&=&
	\frac{(\vect{X}-\vect{c}_{\vect{x}})^T \boldsymbol\Lambda^{-1}
		(\vect{X}-\vect{c}_{\vect{x}})}{2}
	- (\vect{X}-\vect{c}_{\vect{x}})^T \boldsymbol\Lambda^{-1} \vect{W}
		(\vect{Y}-\vect{c}_{\vect{y}}) \nonumber \\
	&&
	- (\vect{Y}-\vect{c}_{\vect{y}})^T \vect{b_y} 
	- (\vect{Z}-\vect{c}_{\vect{z}})^T \vect{b_z}
	- (\vect{Y}-\vect{c}_{\vect{y}})^T \vect{U}(\vect{Z}-\vect{c}_{\vect{z}}),
	\end{eqnarray}
	where $\vect{Y}$ represents the first hidden layer, and $\vect{Z}$
	denotes the second hidden layer. $L$, $M$, $N$ are the dimensionality of $\vect{X}$,
	$\vect{Y}$ and $\vect{Z}$ respectively.  $\Theta:=\left\{\vect{W},
	\vect{U}, \vect{b}_{\vect{y}}, \vect{b}_{\vect{z}} \right\}$ denotes the
	parameters trained for maximizing the loglikelihood. 
	$\mathcal{C}:=\left\{ \vect{c}_{\vect{x}}, \vect{c}_{\vect{y}},
		\vect{c}_{\vect{z}} \right\}$
	represents the offsets for centering the units and is set to the mean
	activation of each unit.  
	$\boldsymbol\Lambda$ is a diagonal matrix with the elements
	$\sigma_i^2$. The probability of any given
	state $(\vect{x}, \vect{y}, \vect{z})$ in the DBM is 
	\begin{eqnarray}
		P\left( \vect{x}, \vect{y}, \vect{z}; \Theta, \mathcal{C} \right) :&=&
		\frac{1}{\mathcal{Z}(\Theta)} 
		\exp\left( -E(\vect{x}, \vect{y}, \vect{z}; \Theta, \mathcal{C})
		\right).
	\end{eqnarray}
	Here $\mathcal{Z}\left( \Theta \right)$ is the partition function
	depended on the parameters of the
	model. Inference in centered GDBM is simple, because the states of the units in
	each layer are independent of the other ones in the same layer given the
	adjacent upper and lower layer. 
	\begin{eqnarray}
		P\left( X_i | \vect{y} \right) &=&
		\mathcal{N}\left( X_i; \vect{w}_{i*} (\vect{y} - \vect{c}_{\vect{y}}) 
			+ c_{x_i}, \sigma_i^2 \right), \\
		P\left( Y_j = 1| \vect{x}, \vect{z} \right) &=&
		f\left( (\vect{x}-\vect{c}_{\vect{x}})^T \vect{w}_{*j} 
		+ \vect{u}_{j*}(\vect{z} - \vect{c}_{\vect{z}}) + b_{y_j} \right), \\
		P\left( Z_k = 1| \vect{y} \right) &=&
		f\left( (\vect{y}-\vect{c}_{\vect{y}})^T \vect{u}_{*k} + b_{z_k} \right), 
	\end{eqnarray}
	where $f\left(\cdot\right)$ is a sigmoid function and
	$\mathcal{N}(\cdot;\mu,\sigma^2)$ denotes a normal distribution with  mean
	$\mu$ and variance $\sigma^2$. $\vect{w}_{i*}$ and $\vect{w}_{*j}$
	denote the $i$th row and the $j$th column of matrix $\vect{W}$.
	$\vect{u}_{j*}$ and $\vect{u}_{*k}$ are defined correspondingly.

	For training a centered GDBM, the objective is to maximize the
	loglikelihood $\hat\ell$,
	of which the partial derivative for each parameter $\theta\in\Theta$ is 
	\begin{eqnarray}
	\frac{ \partial \hat\ell }{ \partial \theta }
	& = &
	\left<\frac{\partial(-E(\vect{x}, \vect{y}, \vect{z})}
			{\partial \theta}\right>_{data} 
	- \left<\frac{\partial(-E(\vect{x}, \vect{y}, \vect{z})}
			{\partial \theta}\right>_{model}, 
	\end{eqnarray}
	where $\left<\cdot\right>_{data}$ and $\left<\cdot\right>_{model}$
	represent the expectation with respect to the data and the model
	distribution, respectively. By using the mean-field approximation, we
	can estimate the data-dependent expectation. 
	And the model-dependent expectation is usually approximated by using
	persistent Markov Chains with Gibbs sampling.
	See~\cite{SalakhutdinovHinton-2009} for details. 
	
	As for the offsets, we adjust them along the training procedure with a
	moving average. To ensure the units are centered from the
	beginning of training, $\vect{c_y}$ and $\vect{c_z}$
	are set to $f\left(\vect{b_y}\right)$ and $f\left(\vect{b_z}\right)$,
	respectively. And $\vect{c}_{\vect{X}}$ is set to the data mean
	$\left<\vect{x}\right>_{data}$.
	In general, we follow the learning
	algorithm in~\cite{MontavonMueller-2012} with several modifications as
	shown in Algorithm~\ref{alg:trainingAlg}.
	\begin{algorithm}[ht] 
	\caption{Training algorithm for centered GDBMs \label{alg:trainingAlg}}
	\begin{algorithmic}[1]
	   \STATE $Initialize~\vect{W}, \vect{U}$ \hfill\big($i.e. ~ w_{ij} \sim
           \mathcal{U}\big[-\sqrt{\frac{6}{L+M}},\sqrt{\frac{6}{L+M}}\big],
	   ~ u_{jk} \sim 
           \mathcal{U}\big[-\sqrt{\frac{6}{M+N}},\sqrt{\frac{6}{M+N}}\big]$\big)
           \STATE $Initialize~\boldsymbol\Lambda$ \hfill
           \big($i.e. ~\sigma_i^2 \sim \mathcal{N}(0.5, 0.01)$\big)
           \STATE $Initialize~\vect{b_y},\vect{b_z}$ \hfill 
           \big($ i.e. ~b_{y_j}, b_{z_k} \sim \mathcal{N}(-4.0, 0.01) $\big) 
           \STATE $Initialize~\vect{c_x},\vect{c_y}, \vect{c_z}$ \hfill
           \big($ i.e. ~\vect{c_x} \gets <\vect{x}>_{data}, 
           \vect{c_y} \gets f(\vect{b_y}), 
           \vect{c_z} \gets f(\vect{b_z}) $ \big)
           \LOOP 
           \STATE $\vect{y}_{model} \gets \vect{c}_{\vect{y}}$
           \FORALL{batches $\vect{x}_{data}$}
              \STATE $\vect{z}_{data} \gets \vect{c}_{\vect{z}}$
              \LOOP
	      \STATE $\vect{y}_{data}~\gets P(\vect{Y}|\vect{x}_{data}, \vect{z}_{data})$
	      \STATE $\vect{z}_{data}~\gets P(\vect{Z}|\vect{y}_{data})$
              \ENDLOOP~until stop criteria is met
              \LOOP
              \STATE $\vect{z}_{model} \sim P(\vect{Z}|\vect{y}_{model})$
              \STATE $\vect{x}_{model} \sim P(\vect{X}|\vect{y}_{model})$
              \STATE $\vect{y}_{model} \sim P(\vect{Y}|\vect{x}_{model}, \vect{z}_{model})$
              \ENDLOOP~until stop criteria is met
              \STATE $\vect{c_{y}} \gets 
              (1-\nu)\cdot\vect{c_y} + \nu\cdot\langle \vect{y}_{data}\rangle$
              \STATE $\vect{c_{z}} \gets 
              (1-\nu)\cdot\vect{c_z} + \nu\cdot\langle \vect{z}_{data}\rangle$
              \STATE $\vect{W} \gets \vect{W} + \eta\big(
              \langle (\vect{x}_{data} - \vect{c_x})\boldsymbol\Lambda^{-1}
                      (\vect{y}_{data} - \vect{c_y})^T \rangle
              - \langle (\vect{x}_{model} - \vect{c_x})\boldsymbol\Lambda^{-1}
                        (\vect{y}_{model} - \vect{c_y})^T \rangle\big)$  
              \STATE $\vect{U} \gets \vect{U} + \eta\big(
              \langle (\vect{y}_{data} - \vect{c_y})
              	(\vect{z}_{data} - \vect{c_z})^T\rangle
              - \langle (\vect{y}_{model} - \vect{c_y})
              	  (\vect{z}_{model} - \vect{c_z})^T\rangle\big)$  
	      \STATE $\sigma_i \gets \sigma_i + \eta_{\sigma}\big(
	      \langle \frac{(x_{i, data}-c_{x_i})^2}{\sigma^3} - 
	      \frac{2(x_{i, data}-c_{x_i}) \vect{w}_{i*} (\vect{y}_{data} - \vect{c_y})}{\sigma^3}\rangle
	      - \langle \frac{(x_{i, model}-c_{x_i})^2}{\sigma^3} - 
	      \frac{2(x_{i, model}-c_{x_i}) \vect{w}_{i*} (\vect{y}_{model} - \vect{c_y})}{\sigma^3}\rangle \big)$
              \STATE $\vect{b_y} \gets \vect{b_y} + \eta\big(
              \langle\vect{y}_{model} \rangle 
              - \langle \vect{y}_{model}\rangle\big) 
              + \nu \vect{W}^T\boldsymbol\Lambda^{-1}(\vect{x}_{data}-\vect{c_x}) 
              + \nu \vect{U}\big(\vect{z}_{data}-\vect{c_z}\big)$
              \STATE $\vect{b_z} \gets \vect{b_z} + \eta\big(
              \langle\vect{z}_{data} \rangle - \langle
              \vect{z}_{model}\rangle\big) 
              + \nu \vect{U}^T\big(\vect{y}_{data}-\vect{c_y}\big)$
              \STATE $\vect{y}_{model} \gets 
              P(\vect{Y}|\vect{x}_{model}, \vect{z}_{model})$
              \ENDFOR
           \ENDLOOP 
	\end{algorithmic}
	\end{algorithm}

\section{Experiments and results} 
\subsection{Learning from natural image patches}
\label{sec:naturalImages}
We applied the centered GDBM to image patches of $32\times32$
pixels\footnote{We generated 60,000 image patches by
randomly taken patches of $32\times32$ pixels from 2,000 natural images. A
subset of size 50,000 was used for training. The rest was used for testing.},
taken randomly from the van Hateren natural image
dataset~\cite{SchaafHateren-1996}. 
The patches were firstly whitened by principal component analysis and reduced
to the dimensionality of $256$ in order to avoid
aliasing~\cite{HyvaerinenKarhunenEtAl-2004}. 

Afterwards, we trained a centered GDBM with 256 visible units and two hidden layers.
There were 900 units in the first hidden layer and $100$ units in the second
hidden layer\footnote{The size of the hidden layers were chosen to get a
	good model of the spontaneous cortical activity as described in
	Section~\ref{sec:spontaneousExp}. The training procedure started with a
	learning rate of 0.03 and a momentum of 0.9, which is annealed to 0.001
	and 0.0, respectively. But the standard deviation $\sigma_i$ have a different 
	learning rate, which is only one-tenth of the other's learning
	rate~\cite{Krizhevsky-2009}. Neither weight decay nor sparse penalty is used
	during training. Mini-batch learning is used with a batch size of
	100. The updating rate for centering parameters was 0.001.
	The training procedure was stopped when the reconstruction error
	stopped decreasing.}. 
Despite the difficulties during training of GDBMs, we found empirically the centered GDBM can be
trained much easier. Without the layer-wise pretraining, the centered GDBM did not
suffer from the issue that the higher layer units are either always
inactive or always active as reported in~\cite{ChoRaikoEtAl-2013}. Since any
centered GDBM can be reparameterized as a normal
GDBM~\cite{MelchiorFischerEtAl-2013}, this may imply that the centering in
GDBM plays an important role in the optimization procedure.

\begin{figure}[ht]
	\centering
	\includegraphics[width=0.7\linewidth]{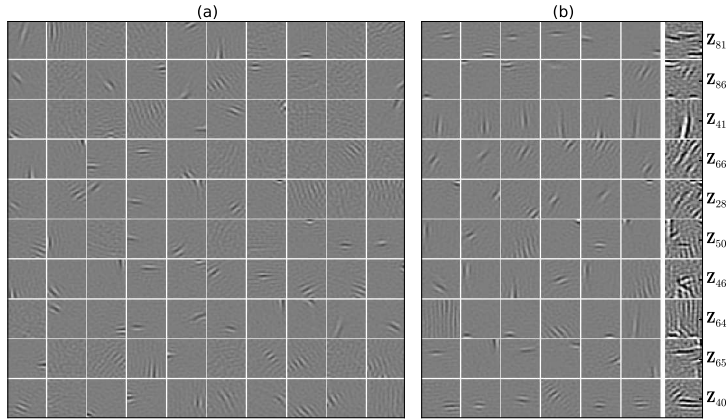}
	\caption{(a) 100 randomly selected first-layer filters
		learned from the natural image. (b) The leading six columns 
		visualize the first-layer filters that have strongest connections
		to one selected second-layer unit. The filters are arranged
		from left to right in descending order by
		the absolute value of their weight to the selected
		unit. The last column depicts the
		weighted sum of the six strongest-connected filters, which
		can be considered as an approximation of the receptive fields of the selected
		second-layer units.}
	\label{fig:gdbmWU}
\end{figure}

After training, the centered GDBM had oriented, Gabor-like filters in the first
hidden layer (Figure~\ref{fig:gdbmWU}a). Most of the units in the second hidden layer
had either strong positive or negative connections to the filters
in the first layer that have similar patterns. As shown in
Figure~\ref{fig:gdbmWU}b, the filters having strong connections to the same
second-layer units either have the similar orientation or the same location. 
The results further suggest that the model learned to encode more complex features
such as contours, angles, and junctions of edges. These results resemble the
properties of the neurons in V1 of visual cortex and imply that 
centered GDBM is a meaningful model for basic receptive field properties in
early cortical visual areas. Despite the
resemblance of these results to those from sparse DBNs~\cite{LeeEkanadhamEtAl-2007}, 
sparse DBNs show worse match to the biological findings
in~\cite{KenetBibitchkovEtAl-2003}. A quantitative comparison between
centered GDBMs and sparse DBNs is still open for future studies. 

\subsection{Comparing with biological experiments}
\label{sec:spontaneousExp}
After successful training of centered GDBM, we quantitatively analyzed the results of the
trained model\footnote{Considering the
	authors of the original paper~\cite{KenetBibitchkovEtAl-2003} only
	presented the results from one hemisphere of a
	selected cat, here we only present the results of one centered GDBM.
	However, all the centered GDBMs trained in our experience showed
consistent results. }
with the methods used in~\cite{KenetBibitchkovEtAl-2003}. For conducting the
measurements, we made two basic assumptions. Firstly, since only the
first-layer filters of centered GDBM present strong orientation preferences, we assumed
these filters correspond to the visual cortical neurons recorded in the
literature. Secondly, the samples of the hidden units collected according to
the model distribution was assumed to be the counterpart of the neuron's
activity.

\subsubsection{Generating orientation maps}
To compare with the original experiments, we firstly generated full-field
gratings as input data and measured the response of the centered GDBM
model. We collected the responses of the first-layer hidden units to
eight orientations from $0^{\circ}$ to $157.5^{\circ}$ with a difference of
$22.5^{\circ}$.

\begin{figure}[ht]
	\centering
	\includegraphics[width=0.65\linewidth]{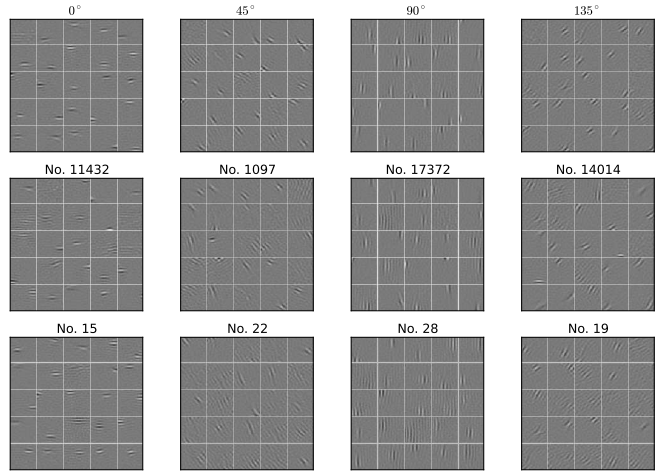}
	\caption{(top) The 25 most active filters in each of the
		single-condition orientation maps. For each map the filters are
		arranged in a descending order of activity level.
		(middle) The same but using the filters in the spontaneous frames
		that are best correlated to the corresponding single-condition
		maps.
		(bottom) The same but using the filters of the best-correlated
		nodes in the self-organizing map.}
	\label{fig:activeFilters}
\end{figure}

The amplitude of the gratings was chosen such that the
average norm of the input stimuli is the same as that of the natural image
patches before whitening.
For each orientation, the grating stimuli of various frequencies and
phases was fed to the model. To be consist with the learning procedure, we used
the mean-field variational approximation to approximate the responses of the
model to each stimulus, which can also be estimated by using the Gibbs sampling
with the visible units clamped to the input. An
illustration of the sampling procedure in given in Figure~\ref{fig:gdbm}(c).
After collecting the responses, the
average response over all the stimuli of each orientation were calculated and
considered to be the model's response to the corresponding orientation. These
activity patterns correspond to the \emph{single-condition orientation maps}
in~\cite{KenetBibitchkovEtAl-2003}. Figure~\ref{fig:activeFilters} (top)
visualizes the most active filters in the single-condition orientation maps of
four selected orientations.
\subsubsection{Generating spontaneous frames}
To simulate spontaneously emerging cortical activity, we sampled the states
of the trained model starting from a random initialization. Neither the visible
nor the hidden unit was clamped to fixed value during sampling.
Figure~\ref{fig:gdbm}(b) shows an illustration of the sampling procedure. This
sampling procedure generated samples from the model's distribution
$P(\vect{Y},\vect{X},\vect{Z})$, which were considered as the expectated states
of the units in the model. This differes from the sampling procedure
in~\cite{ReichertSeriesEtAl-2010}, where the visible units were clamped to
zeroes and the hidden biases were adapted during sampling. The difference is
discussed at length in Section~\ref{sec:discussion}.

In total, there were 100 Markov chains running Gibbs sampling for 2,000 iterations. 
For each Markov chain, the initial states of the first-hidden-layer units were
set to be active with a probability that is equal to the average
$P(\vect{Y}|\vect{x}_{data},\vect{z}_{data})$ over the training data.
By recording the samples every 10 sampling steps, we collected 20,000
samples of $\vect{Y}$ and considered the corresponding probabilities
$P(\vect{Y}|\vect{x}, \vect{z})$ as \emph{spontaneous frames}. Such a
sampling procedure is referred to as a session. In the experiments we
repeated the session for many times with every trained model.

\subsubsection{Correlation between spontaneous frames and orientation maps}
\begin{figure}[ht]
	\centering
	\includegraphics[width=1.0\linewidth]{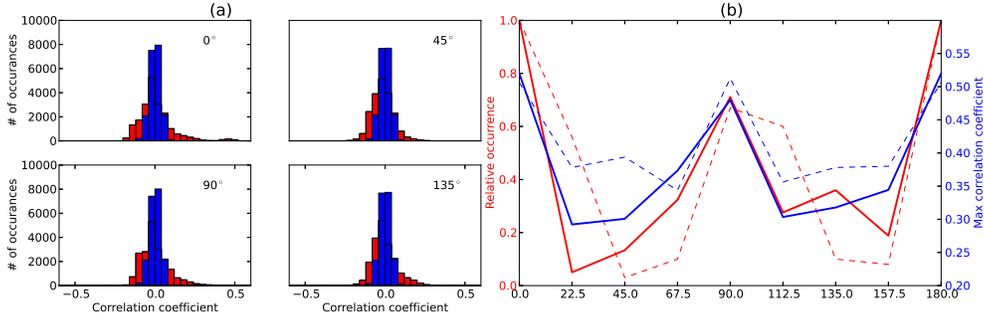}
	\caption{(a) Red, example distribution of the correlation coefficients
	between the spontaneous frames and the four selected single-condition
	orientation maps. Blue, the same, using random generated activity
	patterns with the same average probability. The threshold for
	significant correlation with the significance level of $P
	< 0.01$ is $|0.182|$. (b) The relationship
	between the relative occurrence of the different orientation
	preferences (red) and the maximal correlation coefficient (blue).
	The results from the centered GDBM are plotted by solid lines in
	comparison with the results in~\cite{KenetBibitchkovEtAl-2003} plotted
	by dotted lines. The relative occurrence is calculated relative to the
	occurrence of the horizontal orientation. The figure is adapted to Figure
	3b in~\cite{KenetBibitchkovEtAl-2003}.}
	\label{fig:correlation}
\end{figure}

To establish the similarity between the single-condition orientation maps and
the spontaneous frames, we calculated the spatial correlation coefficients
between them.
Figure~\ref{fig:correlation}(a, red) presents an example of the distribution of
these correlation coefficients for four selected orientations\footnote{The
	following results were collected in a single session, but the results
	are consistent across simulation runs. In all sessions, the similar observation as shown in
	Figure~\ref{fig:gdbmWU}--\ref{fig:correlation} can be observed. The
	only difference is the shape of the curves in
	Figure~\ref{fig:correlation}b might vary in different sessions. However,
	the dominance of the cardinal orientation is always present.}. 
In order to show that these correlations are stronger than expected by chances,
we generated random activity patterns of the first hidden layer with the same
probability as the one used for initializing the Markov chains, and the same
correlation coefficients were calculated with these random activity patterns as
shown in Figure~\ref{fig:correlation}(a, blue). Specifically, the maximal
correlation coefficients is $0.50\pm0.06$ whereas the correlation coefficients
between the spontaneous frames and the random generated patterns seldom reach $0.2$. 
The same observation were made in Figure $2$ in~\cite{KenetBibitchkovEtAl-2003}.

In~\cite{KenetBibitchkovEtAl-2003}, the authors observed that there were more
spontaneous activity patterns corresponding to the cardinal orientations than
to the oblique ones. 
Therefore we further calculated the orientation preference of these spontaneous
frames. By this point, only the spontaneous frames that are significantly correlated
were chosen. As in the biological experiment~\cite{KenetBibitchkovEtAl-2003},
we chose a significance level of $P < 0.01$, resulting in a threshold of
$|0.182|$ and a selection of $18\pm9\%$ spontaneous frames. We then
calculated the orientation preference of these frames by searching the
orientation that is maximally correlated for each frame. Figure~\ref{fig:correlation}(b) plots
the relative occurrences of the different orientation preferences
together with the maximal correlation coefficients. 
The results match those from the cats' visual cortex
in~\cite{KenetBibitchkovEtAl-2003} fairly well, i.e. the spontaneous frames
corresponding to the cardinal orientations emerged more often than those
corresponding to the oblique ones and the former also have larger correlation
coefficients. 

Next we compared the most active filters in the
spontaneous frames with those in the single-condition orientation map.
Figure~\ref{fig:activeFilters} (middle) visualizes these filters for four
spontaneous frames that are best correlated with the selected
single-condition orientation maps. These filters demonstrate similar features
as those in the corresponding orientation maps shown in
Figure~\ref{fig:activeFilters} (top). Thus the result supports the
similarity between the spontaneous frames and the orientation maps in centered GDBMs.

\subsubsection{Learning SOM from the spontaneous frames}
We followed the methods in~\cite{KenetBibitchkovEtAl-2003} and
applied Self-Organizing Map (SOM) algorithm~\cite{Kohonen-2001} to the spontaneous frames in order
to study the intrinsic structures of these spontaneous activities. We trained a SOM
on the 20,000 spontaneous frames collected from a single session. The SOM projects
the spontaneous frames onto 40 nodes that were arranged on a 1-D circle.
See~\cite{KenetBibitchkovEtAl-2003} for details of training. After training the
SOM, we examined the correlation between the weight vectors of the 40 nodes and
the single-condition orientation maps. Figure~\ref{fig:activeFilters} (bottom)
illustrates the most active filters in the weight vectors of four nodes that
are best correlated with the selected single-condition orientation maps. The
remarkable resemblance between these filters and those in the single-condition orientation
maps suggests that the spontaneous frames encompass several states of the
hidden variables in the first layer, which resemble the model's representation
of orientation.

\section{Discussion}
\label{sec:discussion}
In this work, we present a variant of DBMs, centered Gaussian-binary
deep Boltzmann machines (GDBM) for modeling spontaneous activity in visual cortex. 
We show empirically that the proposed centered GDBM does not require the layer-wise pretraining procedure by
virtue of centering the units' activation. An intuitive explanation for the success of centering
is that centering prevents the units from being either always active or
inactive and thus forming a bias for the other units. Because the centering
offsets keep the output of each unit to zero unless its activation differs from
its mean activation. Formally, the authors in~\cite{MontavonMueller-2012} show that the
centering produces improved learning conditions.
Nevertheless, the authors show that a centered DBM can be
reparameterized as a normal DBM with the same probability distribution
in~\cite{MelchiorFischerEtAl-2013}. Therefore, centered GDBM can be considered
as a GDBM with better optimization conditions. In other word, the results
presented in this work can also be expected in a GDBM, despite its optimization
difficulties. 

The results of modeling natural image patches with centered GDBM in this work
suggest centering helps to overcome the commonly observed difficulties during training. In
addition, we also trained a centered GDBM on the Olivetti face
dataset~\cite{SamariaHarter-1994} following the same setting as
in~\cite{ChoRaikoEtAl-2013}. We achieved a
reconstruction error on the test set of $41.6\pm0.40$ compared to the
result of about $40$ in~\cite{ChoRaikoEtAl-2013}. 
Although the results are not sufficient to claim the
superiority of centered GDBMs over the proposed training algorithm
in~\cite{ChoRaikoEtAl-2013}, we argue that a centered GDBM is an
alternative to GDBM.  

Our main contribution is to consider centered GDBMs as a model for the
early cortical visual areas and to reproduce the findings of the spontaneous cortical
activity in\cite{KenetBibitchkovEtAl-2003}. Compared to previous
work~\cite{ReichertSeriesEtAl-2010} of modeling cortical activity with DBMs,
we extend the model from binary data to continuous data with the proposed
centered GDBMs. This extension makes it possible to use centered GDBMs to model
other cortical activities besides vision. 

For modeling spontaneous activity, we've also tested other models, i.e.
GRBMs and DBN. None of them can match the results
in~\cite{KenetBibitchkovEtAl-2003} as well as centered GDBMs. 
The correlation between the spontaneous frames and the orientation maps are
significantly less than in centered GDBMs, where $18\pm9\%$ of the frames are
significantly correlated to the maps compared to $2\%$ or less in other
models. A possible explanation is that centered GDBMs is the only model using
both top-down and bottom-up interactions during inference. In comparison, GRBMs
and DBN only use either the bottom-up or the top-down information. On one hand,
this suggests that the observed
spontaneous activity is a result of interactions between 
incoming stimulation and feedback from higher areas. On the other hand, our
results also predicts the states of spontaneous activity found
in~\cite{KenetBibitchkovEtAl-2003} are the result of the interactions within the
early cortical visual areas. Because a centered GDBM with two hidden layers is enough
to faithfully reproduce the reported results.

In~\cite{ReichertSeriesEtAl-2010}, the authors clamped the visible units to
zero in order to model profound visual impairment or blindness. This is
equivalent to clamping the visible units of a centered GDBM to the centering offsets, which
leads to zero inputs to the first hidden layer from bottom-up. However,
in this work we sampled the visible units freely during generating spontaneous
frames. The main reason is that spontaneous cortical activity is the ongoing activity in
the absence of intentional sensory input, which does not exclude incoming
stimuli~\cite{ArieliSterkinEtAl-1996}. The authors also reported no difference
in their findings when a uniform grey screen is used instead of having the room
darkened~\cite{KenetBibitchkovEtAl-2003}. From the model's persepctive, our
sampling procedure is supposed to approximate samples from the model's prior
distribution $P(\vect{Y})$, which are the expectated states of the
first-hidden-layer units without any knowledge of bottom-up or top-down
information. As a result, our results suggest that the spontaneous activity
patterns appear to be an expectation of internal states in brain and indicate
that a brain might learn to generate such expectations as a generative model.
Moreover, we observed that the correlation between the spontaneous frames and
the orientation maps disappeared when we clamped either the visible or the
second hidden layer units to the centering offsets during generating the
spontaneous frames. This supports our prediction that generating the observed
spontaneous activity needs both incoming stimulation and feedback from higher areas.

\section{Conclusion}
We present centered GDBMs by applying the centering to GDBMs and
used the centered GDBM as a model of spontaneous activity in early cortical visual areas.
Our work extend the previous work of using DBM for modeling cortical activities
to continuous data. The results demonstrate that a centered GDBM is a
meaningful model approach for basic receptive field properties and the
emergence of spontaneous activity patterns in early cortical visual area and
has the potential to give further insights to 
spontaneous activity in brain. Our work also show empirically 
centered GDBMs can be properly trained without layer-wise pretraining. 

\subsubsection*{Acknowledgments}
We would like to thank Jan Melchior for helpful comments. This work is funded
by a grant from the German Research Foundation (Deutsche
Forschungsgemeinschaft, DFG) to L. Wiskott (SFB 874, TP B3) and D. Jancke
(SFB-874, TP A2).

\newpage
\bibliographystyle{unsrt}
\subsubsection*{References}
\begin{footnotesize}
\bibliography{spontaneousCortexActivity}
\end{footnotesize}
\end{document}